%% file: main.tex
\documentclass{article}

\usepackage{microtype}
\usepackage{graphicx}
\usepackage{subfigure}
\usepackage{booktabs} 

\usepackage{hyperref}

\usepackage[accepted]{icml2024}

\usepackage{amsmath}
\usepackage{amssymb}
\usepackage{mathtools}
\usepackage{amsthm}

\usepackage[capitalize,noabbrev]{cleveref}

\theoremstyle{plain}

\theoremstyle{definition}

\theoremstyle{remark}

\usepackage[textsize=tiny]{todonotes}

\input{setup}

\icmltitlerunning{In-Context Learning with Topological Information for Knowledge Graph Completion}

\begin{document}

\twocolumn[

\icmltitle{In-Context Learning with Topological Information for \\LLM-Based Knowledge Graph Completion}



\icmlsetsymbol{equal}{*}

\begin{icmlauthorlist}
\icmlauthor{Udari Madhushani Sehwag}{jpmc,equal}
\icmlauthor{Kassiani Papasotiriou}{jpmc,equal}
\icmlauthor{Jared Vann}{jpmc}
\icmlauthor{Sumitra Ganesh}{jpmc}
\end{icmlauthorlist}

\icmlaffiliation{jpmc}{JPMorgan AI Research}

\icmlcorrespondingauthor{Udari Sehwag}{udari.madhushani@jpmorgan.com}

\icmlkeywords{Machine Learning, ICML}

\vskip 0.3in
]



\printAffiliationsAndNotice{\icmlEqualContribution} 

\begin{abstract}
\input{sec/abstract}
\end{abstract}


\section{Introduction}
\input{sec/intro}

\section{Related work}
\input{sec/related_work}

\section{Problem formulation}
\input{sec/problem_formulation}

\section{Methodology}
\input{sec/methodology}

\section{Results}
\input{sec/results}

\section{Discussion and conclusions}
\input{sec/conclusion}

\section{Limitations}
\input{sec/limitations}


\section*{Disclaimer}
This paper was prepared for informational purposes in part by the Artificial Intelligence Research group of JPMorgan Chase \& Co and its affiliates (“JP Morgan”),
and is not a product of the Research Department of JP Morgan. JP Morgan makes no representation and warranty whatsoever and disclaims all liability, for the completeness, accuracy or reliability of the information contained herein. This document is not intended
as investment research or investment advice, or a recommendation, offer or solicitation for
the purchase or sale of any security, financial instrument, financial product or service, or
to be used in any way for evaluating the merits of participating in any transaction, and
shall not constitute a solicitation under any jurisdiction or to any person, if such solicitation
under such jurisdiction or to such person would be unlawful.

\bibliography{main}
\bibliographystyle{icml2024}

\newpage
\appendix
\onecolumn
\section{Prompts used for results generation}
\input{appendix/prompts}

\section{Paths identified in the ontology}
\input{appendix/ontology_paths}

\end{document}

%% file: setup.tex
\usepackage[utf8]{inputenc} 
\usepackage[T1]{fontenc}    
\usepackage{hyperref}       
\usepackage{url}            
\usepackage{booktabs}       
\usepackage{amsfonts}       
\usepackage{nicefrac}       
\usepackage{microtype}      
\usepackage{xcolor}         

\usepackage{lipsum}
\usepackage{amsmath}
\usepackage{graphicx}
\usepackage{spverbatim}
\usepackage{multirow}
\usepackage{tabularx}
\usepackage{enumitem}

\usepackage{url}
\usepackage{hyperref}
\definecolor{SkyBlue}{HTML}{0077B6}
\hypersetup{
    colorlinks=true,
    bookmarksnumbered,
    linkcolor={blue!80},
    citecolor={SkyBlue},
    urlcolor={blue!80},
}

\usepackage{listings}
\newcommand{\ra}{{$\rightarrow$ }}

%% file: sec/abstract.tex
Knowledge graphs (KGs) are crucial for representing and reasoning over structured information, supporting a wide range of applications such as information retrieval, question answering, and decision-making. However, their effectiveness is often hindered by incompleteness, limiting their potential for real-world impact. While knowledge graph completion (KGC) has been extensively studied in the literature, recent advances in generative AI models, particularly large language models (LLMs), have introduced new opportunities for innovation. In-context learning has recently emerged as a promising approach for leveraging pretrained knowledge of LLMs across a range of natural language processing tasks and has been widely adopted in both academia and industry. However, how to utilize in-context learning for effective KGC remains relatively underexplored.  We develop a novel method that incorporates topological information through in-context learning to enhance KGC performance. By integrating ontological knowledge and graph structure into the context of LLMs, our approach achieves strong performance in the transductive setting i.e., nodes in the test graph dataset are present in the training graph dataset. Furthermore, we apply our approach to KGC in the more challenging inductive setting, i.e., nodes in the training graph dataset and test graph dataset are disjoint, leveraging the ontology to infer useful information about missing nodes which serve as contextual cues for the LLM during inference. Our method demonstrates superior performance compared to baselines on the ILPC-small and ILPC-large datasets.

%% file: sec/intro.tex
Knowledge graphs have emerged as a powerful framework for representing and reasoning over large-scale structured knowledge, with applications spanning question answering 
 \cite{song2023advancements,yani2021challenges}, reccomednation systems \cite{chicaiza2021comprehensive} and decision making \cite{10.1145/3447772}. The effectiveness of knowledge graphs in these domains relies heavily on their completeness and accuracy. However, constructing and maintaining comprehensive knowledge graphs is a challenging task, often requiring extensive manual effort and domain expertise \cite{kgretrospective}. To address this challenge, researchers have explored automated methods for Knowledge Graph Completion (KGC) \cite{shen2022comprehensive, chen2020knowledge}. These methods leverage information extracted from various sources, such as text corpora \cite{wang2021kepler, shi2018open}, web pages \cite{dong2014knowledge, mitchell2018never}, and databases \cite{zou2014gstore}. These methods typically employ natural language processing techniques, such as named entity recognition and relation extraction \cite{9039685, pawar2017relation}, to identify new facts, integrate them into the existing knowledge graphs and identifying new relations between nodes that exist in the graph. The rapid growth of unstructured data has introduced new challenges in automated graph extension methods.

Recent advances in generative AI, particularly the emergence of Large Language Models (LLMs) have opened up new possibilities for extending knowledge graphs \cite{Pan_2024}. Modern LLMs demonstrate impressive capabilities in understanding large texts \cite{geminiteam2024gemini}, generating natural language as well as reasoning \cite{openai2024gpt4} over complex information. These models are trained on vast amounts of unstructured data, allowing them to capture rich semantic knowledge and develop a deep understanding of various domains. Pretrained knowledge of LLMs and their reasoning capabilities can be harnessed to infer connections between nodes and relations predicting missing information in graphs. links in the graph, making LLMs well suited for task of KGC. 

In this paper, we propose a novel approach that leverages graph topological information through in-context learning to enhance the KGC performance of LLMs. Our method is based on the intuition that the extensive knowledge embedded in pretrained language models can provide valuable insights and predictions for missing information in knowledge graphs.  We introduce a two-step process: first, we construct an ontology from the knowledge graph using the LLM's domain understanding, capturing the types of nodes and relationships in the graph. By combining this ontological structure with the graph's topology and employing chain-of-thought (CoT) reasoning, we provide the LLM with context to make more informed predictions. Secondly, our algorithm leverages structured information from the graph, utilizing overlapping nodes between the missing knowledge triplets and the existing graph triplets, combined with the ontology, to generate candidate solutions for the missing information. Additionally, we consider alternative paths between the existing nodes and potential candidate nodes, thereby exploiting the complex topological structure of the graph. This comprehensive use of the graph's topological structure and the LLM's predictive capabilities leads to our method performing significantly better than the state-of-the-art baseline approaches.


Our contributions are as follows: (1) We propose a generative ontology creation method using LLMs to derive ontologies from raw knowledge graph data, capturing the types of nodes and relationships in the graph. (2) We leverage the generated ontology and the graph's topological information, including paths between nodes, to enhance link prediction. (3) By utilizing the ontology to identify candidate solutions for missing triplets and employing the LLM to select the correct solution, we improve KGC performance in both transductive and inductive settings. Importantly, our method requires no additional training, highlighting its efficiency and immediate applicability.

%% file: sec/related_work.tex
There exists a substantial body of work at the intersection of LLMs and knowledge graphs, extensively reviewed in \cite{Pan_2024}. This section provides a detailed exploration of relevant literature.

\paragraph{Knowledge graph completion datasets}
Some of the popular knowledge graph completion datasets include Freebase, a comprehensive knowledge base integrated into Google's Knowledge Graph, from which FB15k \cite{bordes2013translating} and FB15k-237 \cite{toutanova-etal-2015-representing} datasets are derived. WN18RR \cite{dettmers2018convolutional} is a subset of WordNet widely used for link prediction models. ILPC (Inductive Link Prediction Challenge) datasets \cite{Galkin2022} are also significant, featuring both small and large datasets designed for inductive reasoning tasks. Other domain-specific datasets include MEDCIN \cite{medcin} covering biomedical entities and GeoNames \cite{geonames} focusing on geographical entities. These datasets collectively serve as crucial benchmarks for evaluating various aspects of knowledge graph completion models.


\paragraph{Link prediction in knowledge graphs}

One group of methods for link prediction in knowledge graphs involves the use of vector embeddings. \cite{simplE} makes use of background knowledge when creating the embeddings. \cite{ZHANG2021106564} embeds head and tail entities into time and frequency domain spaces respectfully. \cite{Zhang_Cai_Zhang_Wang_2020} shows an embedding scheme based on entity type hierarchies. Another family of methods is the use of trained deep neural network models. \cite{NEURIPS2018_53f0d7c5}, \cite{node_cooccurrence} and \cite{MOHAMED202390} demonstrate the use of GNNs for link prediction in graphs and knowledge graphs. \cite{neelakantan2015compositional} uses RNNs for graph completion. Other methods focus on 'lifelong-learning' and the continual updating of knowledge graphs based on new information \cite{Mazumder_2019}.

\paragraph{LMs for link prediction and ontology creation }
Approaches such as BERTRL \cite{zha2022inductive} and KGT5 \cite{saxena2022sequence} treat each triplet in the knowledge graph as a textual sequence, refining models based on these sequences. These methods leverage information from language model parameters but do not explicitly integrate information extracted from knowledge graph during link prediction. In contrast, frameworks like Better Together \cite{chepurova2023better} and KGT5-context \cite{kochsiek2023friendly} incorporate node neighborhoods directly within the context of generative language models. Additionally, recent studies have explored the capability of large language models (LLMs) to create coherent ontologies. LLMs4OL \cite{giglou2023llms4ol} systematically evaluated various LLMs, demonstrating that models fine-tuned for specific tasks consistently outperformed zero-shot methods. In another approach, Kommineni et al. \cite{kommineni2024human} used LLMs to generate "competency questions," which were employed to develop ontologies for knowledge graphs.

%% file: sec/problem_formulation.tex
In this section, we introduce mathematical notations and formally define the problem of KGC using LLMs incoporating topological information.

\subsection{Knowledge graph with ontology}


Let $\mathcal{O}$ be an ontology and let $\mathcal{G}$ be the corresponding knowledge graph. 

Ontology $\mathcal{O}$ can be defined as $\mathcal{O} = (\mathcal{C}, \mathcal{R}, \mathcal{E})$, where:
\setlist{nolistsep}
\begin{itemize}[noitemsep]
    \item $\mathcal{C}$ consists of ontology nodes, node categories of the nodes in the graph,
    \item $\mathcal{R}$ is the set of relations,
    \item $\mathcal{E}$ consists of unique triplets  $(c_i, r, c_j)$ where $c_i, c_j \in C$ and $r \in \mathcal{R}$.
\end{itemize}

Graph $\mathcal{G}$ can be defined as $\mathcal{G} = (\mathcal{V}, \mathcal{R}, \mathcal{T})$, where:
\setlist{nolistsep}
\begin{itemize}[noitemsep]
    \item $\mathcal{V}$ is the set of nodes, where each node $v_i \in \mathcal{V}$ is associated with at least one category $c_{v_i} \in \mathcal{C}$.
    \item $\mathcal{R}$ is the set of relations,
    \item $\mathcal{T}$ consists of triplets formed according to the ontology triplets $\mathcal{E}$. For nodes $v_i$ of category $c_{v_i}$ and $v_j$ of category $c_{v_j}$,  $(v_i,r, v_j) \in \mathcal{T}$ such that $(c_{v_i}, r, c_{v_j}) \in \mathcal{E}$.
\end{itemize}

\subsection{Knowledge graph completion}

Knowledge Graph Completion (KGC) is the task of inferring missing information in a knowledge graph. Given a training knowledge graph $\mathcal{G}_{\text{train}} = (\mathcal{V}, \mathcal{R}, \mathcal{T}_\text{train})$ with some missing triplets $(v_i, r, v_j) \in \mathcal{T}_\text{inference}$, the objectives are to: (1) predict the missing relation $r \in \mathcal{R}$ between two existing entities $v_i, v_j \in \mathcal{V}$, i.e., $(v_i, ? , v_j)$; (2) predict the missing tail entity $v_j \in \mathcal{V}$ given the head entity $v_i \in \mathcal{V}$ and the relation $r \in \mathcal{R}$, i.e., $(v_i, r ,?)$; and (3) predict the missing head entity $v_i \in \mathcal{V}$ given the relation $r \in \mathcal{R}$ and the tail entity $v_j \in \mathcal{V}$, i.e., $(?, r, v_j)$. Typically in the literature \cite{galkin2021nodepiece, chepurova2023better} the problem of predicting the head node is converted to a tail node prediction problem by using the inverse relationship, i.e., $(v_j, r^{-1}, v_i)$. In our experiments, we focus on node prediction rather than relation prediction. We consider the graphs where the relations in the test graph dataset is a subset of the relations in the trainning graph dataset.

\paragraph{Transductive vs inductive link prediction}
KGC tasks can vary based on the type of knowledge graph: inductive or transductive. In a transductive setting, the task is formulated as follows: given a training knowledge graph $\mathcal{G}_{\text{train}} = (\mathcal{V}, \mathcal{R}, \mathcal{T}_\text{train})$, train the model to predict on the inference triplets $(v_i, r, v_j) \in \mathcal{T}_\text{inference}$, where $v_i, v_j \in \mathcal{V}$, i.e., nodes in inference triplets are present in the training graph. In inductive KGC, the model is trained on $\mathcal{G}_{\text{train}} = (\mathcal{V}_\text{train}, \mathcal{R}, \mathcal{T}_\text{train})$ and predicts on the inference triplets $(v_i, r, v_j) \in \mathcal{T}_\text{inference}$, where $v_i, v_j \not\in \mathcal{V}_\text{train}$. By nature KGC in the inductive setting is more challenging compared to the transductive setting.

%% file: sec/methodology.tex
In this section, we outline our approach to KGC encompassing both transductive and inductive settings. 


\subsection{Generating ontology}
To generate the ontology $\mathcal{O} = (\mathcal{C}, \mathcal{R}, \mathcal{E})$ of a knowledge graph $\mathcal{G} = (\mathcal{V}, \mathcal{R}, \mathcal{T})$, for each relation $r \in \mathcal{R}$, we create two node sets $V_i$ and $V_j$ of length $n$ that contain head and tail nodes $(v_i, v_j) \in \mathcal{V}$ connected by $r$. We then prmopt GPT-4 model to predict a head category $c_{i}$ and a tail category $c_{j}$ for entities in $V_i$ and $V_j$ respectively, given the relation $r$. Note that this results in $c_i = c_{v_i}, \forall v_i \in \mathcal{V}_i$ and $c_j = c_{v_j}, \forall v_j \in \mathcal{V}_j$. Naively prompting the LLM to create an ontology by providing node pairs corresponds to a given relation to inconsistencies. GPT-4 model can assign synonyms of node categories to nodes of same class connected to different relations. For example GPT-4 can assign node class label \textit{film} to head node of the triplet (\textit{cruel intentions}, \textit{film cast member actor}, \textit{Alaina Reed Hall}) and assign a node class label \textit{movie} to head node of the triplet (\textit{Dear America: Letters from home Vietnam}, \textit{directed by}, \textit{Bill Couturié}). To improve the quality of the created ontology, we adopt an iterative generation approach that incorporates the previously created sub-ontology at each step. This ensures consistency in node class assignments across similar nodes. Furthermore, we add the triplet $(c_{v_i}, r, c_{v_j})$ to the set $\mathcal{E}$ to associate each relation $r \in \mathcal{R}$ with exactly one pair of node categories $(c_{v_i}, c_{v_j})$, maintaining a well-structured ontology.

\subsection{Link prediction using topology of the ontology}

For a triplet $(v_i, r, v_j) \in \mathcal{T}_\text{test}$ with a missing tail $v_j$ (or missing head $v_i$), we use the generated ontology $\mathcal{O}$ to infer the category $c_{v_j}$ of $v_j$, based on the relation $r$ and the category $c_{v_i}$ of the head. We provide this inference as a hint to the LLM. Consider the example triplet $(\textit{Miles Davis}, \textit{died In}, ?).$
There are multiple choices for the answer: it can be a city, country, hospital, etc. In our method   we find the ontology triplet $(\textit{musician}, \textit{died In}, \textit{country})$ and let the LLM know that the answer is of type $\textit{country}$. Additionally, we compute ontology paths between $c_{v_i}$ and $c_{v_j}$, providing these as context to the LLM. For instance, alternative paths between $\textit{musician}$ and $\textit{country}$ could include $(\textit{musician}) \rightarrow \textit{part Of Band} \rightarrow (\textit{band}) \rightarrow \textit{conceived In Country} \rightarrow (\textit{country})$.

\subsection{Link prediction using topology of the graph}
Recall that in transdictive setting nodes in the test graph dataset are present in the training graph dataset. For a triplet $(v_i, r, v_j) \in \mathcal{T}_\text{inference}$ with a missing tail $v_j$, we infer its category $c_{v_j}$ using the ontology $\mathcal{O}$. Next, using $\mathcal{V}_{\text{train}}$ and the category $c_{v_j}$, we create candidate solutions $v_{\text{candidate}_i} \in \mathcal{V}_{\text{train}}$ that are of category $c_{v_j}$ and prompt the LLM to use the list of candidates as an hint for predicting the missing node. However, the number of candidate nodes we extract from $\mathcal{V}_{\text{train}}$ using the information from ontology can be very large and exceed the limit of the context window. In order to address this problem we employ a strategy which maks multiple LLM calls with sublists of the candidate nodes. Each call tasks the model with selecting the answer that most is most likely to be the missing node. The winning candidates from these calls are subsequently aggregated and provided in a final LLM call as hints to predict the ultimate solution.

\subsection{Chain-of-thought style reasoning}
In our methodology, we employ chain-of-thought (CoT) reasoning to guide the LLM in making informed predictions about missing nodes in the knowledge graph. CoT reasoning involves providing the LLM with a series of intermediate reasoning steps or questions that help break down the problem and lead to the final answer. By incorporating CoT prompts, we encourage the LLM to consider relevant information from the ontology and graph structure, enhancing its ability to make accurate predictions. The CoT prompts are designed to prompt the LLM to reason about the potential missing node based on the available information, such as the ontology paths, graph paths specific to the given triplet, and the ontology hint about the missing node type. This approach helps the LLM make logical connections between the available information and the missing node, ultimately leading to improved performance in knowledge graph completion tasks.

%% file: sec/results.tex

In this section, we present our experimental setup, including the datasets used, baselines compared, and evaluation metrics employed. We then discuss the results obtained from our experiments and provide a comprehensive analysis of our findings, highlighting the key insights and implications.
\begin{table*}
\centering
\caption{Dataset details for ILPC-small and ILPC-large.}
\label{tab:dataset_stat}
\renewcommand{\arraystretch}{1.0}
\resizebox{0.9\linewidth}{!}{
 \begin{tabular}{lccccccc} \toprule
\multirow{2}{*}{\textbf{Dataset}} & \multicolumn{3}{c}{\textbf{ILPC-small}} & & \multicolumn{3}{c}{\textbf{ILPC-large}} \\ \cmidrule{2-4} \cmidrule{6-8}
 & \textbf{\# nodes} & \textbf{\# relations} &  \textbf{\# triplets} & &\textbf{\# nodes} & \textbf{\# relations} & \textbf{\# triplets} \\ \midrule
Inductive training graph & 10,230 & 96 & 78,616 & & 46,626 & 130 & 202,446 \\
Transductive training graph & 6,653 & 96 & 20,960 & & 29,246 & 130 & 77,044 \\
Ontology graph & 36 & 96 & 96 & & 42 & 130 & 130  \\
Inference test graph & 6,653 & 96 & 2,902 & & 29,246 & 130 & 10,184\\
\bottomrule
\end{tabular}}
\end{table*}

\begin{table*}[ht]
\centering
\caption{Hits@k results on ILPC-small, and ILPC-large in transductive setting.}
\label{tab:results_transductive}
\resizebox{0.9\linewidth}{!}{
\begin{tabular}{lccccccc}
\toprule
\multirow{2}{*}{\textbf{Method}} & \multicolumn{3}{c}{\textbf{ILPC-small}} & & \multicolumn{3}{c}{\textbf{ILPC-large}} \\ \cmidrule{2-4} \cmidrule{6-8}
& \textbf{H@1} & \textbf{H@3} & \textbf{H@10} & &\textbf{H@1} & \textbf{H@3} & \textbf{H@10} \\ \midrule
GPT-4 & 0.132 & 0.208 & 0.289 & & 0.146 &  0.204 &  0.253 \\
GPT-4 + neighbors \cite{chepurova2023better} & 0.122 & 0.202 & 0.288 & & 0.154 & 0.208 & 0.273 \\
\midrule
GPT-4 + candidate solutions  & 0.172 & 0.233 & 0.319 & & 0.177 & 0.246 & 0.292 \\
GPT-4 + candidate solutions + ontology  & 0.173 &  0.234 & 0.318 & & 0.176 & 0.245 & 0.292 \\
GPT-4 + candidate solutions + ontology paths  & \bf{0.174} & \bf{0.237} & \bf{0.322} & &  \bf{0.178} & \bf{0.251}  & \bf{0.300} \\
GPT-4 + candidate solutions + ontology + ontology paths  & \bf{0.174} & 0.236 & 0.317 & & 0.177 & 0.249 & 0.297 \\
\bottomrule
\end{tabular}}
\end{table*}

\begin{table*}[t]
\centering
\tiny
\caption{Hits@k results on ILPC-small, and ILPC-large in inductive setting.}
\label{tab:results_inductive}
\resizebox{0.9\linewidth}{!}{
\begin{tabular}{lccccccc}
\toprule
\multirow{2}{*}{\textbf{Method}} & \multicolumn{3}{c}{\textbf{ILPC-small}} & & \multicolumn{3}{c}{\textbf{ILPC-large}} \\ \cmidrule{2-4} \cmidrule{6-8}
& \textbf{H@1} & \textbf{H@3} & \textbf{H@10} & &\textbf{H@1} & \textbf{H@3} & \textbf{H@10} \\ \midrule
IndNodePiece \cite{galkin2022open} & 0.007 & 0.022 & 0.092 & & 0.037 & 0.054 & 0.125 \\
IndNodePieceGNN \cite{galkin2022open} & 0.076 & 0.140 & 0.251 & & 0.032 & 0.073 & 0.146 \\ \midrule
GPT-4 + ontology  & 0.134 & 0.207 & 0.288 &  & 0.150 & 0.202  & 0.247\\
GPT-4 + ontology paths  & 0.135 & 0.\bf{210} & \bf{0.290} &  & 0.150 & \bf{0.208}  & \bf{0.256}\\
GPT-4 + ontology + ontology paths  & \bf{0.138} & 0.208 & 0.289 & & \bf{0.152} & 0.205 & 0.252 \\
 \bottomrule 
\end{tabular}}
\end{table*}

\begin{table*}
    \centering
    \caption{Hits@k results on ILPC-small dataset with different numbers of candidate nodes in the context. The numbers given in the table are for the case where GPT-4 is prompted to choose one candidate node from each sub candidate node list. Numbers in parentheses represent the case where multiple candidate nodes are selected from each sublist, totaling 100 candidate nodes.}
    \resizebox{0.9\linewidth}{!}{
    \begin{tabular}{lccc} 
        \toprule
        \multirow{2}{*}{\textbf{Method}} & \multicolumn{3}{c}{\textbf{ILPC Small}} \\ \cmidrule{2-4}
         & \textbf{H@1} & \textbf{H@3} & \textbf{H@10} \\ \midrule
        GPT-4 + candidate solution & 0.172 (0.196) & 0.233 (0.305) & 0.319 (0.391) \\
        GPT-4 + candidate solutions + ontology & 0.173 (0.187) & 0.274 (0.290) & 0.318 (0.383) \\
        GPT-4 + candidate solutions + ontology paths  & 0.174 (0.198) & 0.237 (0.295) & 0.322 (0.389) \\
        GPT-4 + candidate solutions + ontology + ontology paths & 0.174 (0.196) & 0.236 (0.294) & 0.317 (0.386) \\
    \bottomrule
    \end{tabular}}
    \label{tab:candidate_node}
\end{table*}

\begin{table}
    \centering
    \caption{Number of accurate head node predictions obtained by direct head node prediction compared to predicting tail node with inverse relation in ILPC-small dataset.}
    \resizebox{\linewidth}{!}{
    \begin{tabular}{lccc} 
        \toprule
        \multirow{2}{*}{\textbf{Method}} & \multicolumn{3}{c}{\textbf{ILPC Small}} \\ \cmidrule{2-4}
         & \textbf{H@1} & \textbf{H@3} & \textbf{H@10} \\ \midrule
         GPT-4 & 4 & 9 & 12 \\
         GPT-4 + ontology & 4 & 10 & 20 \\
         GPT-4 + ontology + paths & 6 & 19 & 20 \\
         GPT-4 + ontology + ontology paths & 2 & 7 & 19 \\ \bottomrule
    \end{tabular}}
    \label{tab:direct_inverse}
\end{table}

\paragraph{Datasets and ontology}

We utilized the small and large datasets from the Inductive Link Prediction Challenge (ILPC) 2022 \cite{Galkin2022}. The ILPC datasets are specifically designed for inductive link prediction, meaning they contain disjoint inference graphs with new, unseen entities. Both the small and large ILPC datasets consist of three subsets: inductive training graph dataset, transductive training graph dataset and inference test set. All the evaluations were done on the inference test set.

To create the ontology, we first combined the inductive and transductive training graphs, resulting in a graph with approximately 900k triplets. Following the methodology described in Section 4.1, for each relation in the resulting graph, we sampled 50 examples that are connected by this relation and prompted OpenAI GPT-4 to create two ontology categories that best describe the head and tail examples. We allowed the LLM the option to select already predicted categories if it found them appropriate. We then classified all triplets connected by the relation using the LLM-predicted head and tail ontology categories. We performed a post verification step which validated that each relation is associated with only one pair of node categories. Details of the datasets and ontology are given in Table \ref{tab:dataset_stat}

\paragraph{Experiment Settings}
During inference, for each triplet in the test set, we used GPT-4 to predict the head and the tail under the following conditions: (1) no context, (2) hint derived from the topology of the ontology, (3) hints derived from the topology of both ontology and graph.

\textit{No context}. In no-context setting we provide GPT-4 with the triplet with missing node and promt it to directly predict the missing node. The results for this setting in reported as GPT-4 + vanilla in Table \ref{tab:results_transductive}.

\textit{Ontology hints}. In this setting we provide supporting hints constructed from the topology of the ontology. We consider 4 experimental settings under this method. In \textit{GPT-4 + ontology} we provide GPT-4 with the category of the missing node infered from the ontology triplet that corresponds to the given relation relation. Ontology paths are the paths that exists between available node and missing node categories in the ontology, except the given relation. In \textit{GPT-4 + ontology paths} we provide the path details an additional hint and in \textit{GPT-4 + ontology + ontology paths} we provide both category of the missing node and alternate ontology paths as hints.

\textit{Ontology and graph hints}.  In this setting, we leverage the topology of the graph to generate candidate solutions for the missing node and provide them as hints to GPT-4. We consider four experimental settings under this method. In \textit{GPT-4 + candidate solutions}, we infer the category of the missing node using the ontology and provide GPT-4 with a list of candidate nodes from the training graph that belong to the same category. In \textit{GPT-4 + candidate solutions + ontology}, we provide both the candidate solutions and the ontology-inferred category of the missing node as hints. \textit{GPT-4 + candidate solutions + ontology paths} extends the previous setting by including the alternate paths between the available node and the missing node categories in the ontology as additional hints. Finally, in \textit{GPT-4 + candidate solutions + ontology + ontology paths}, we provide GPT-4 with the candidate solutions, the ontology-inferred category of the missing node, and the alternate ontology paths as comprehensive hints to guide the prediction of the missing node.

\textit{Candidate node selection.}  To generate candidate nodes for the missing node in the transductive setting, we leverage the ontology and the graph structure. First, we infer the category of the missing node from the ontology based on the given relation. Then, we identify all the nodes in the graph that belong to the inferred category and are connected to the available node through the given relation. However, due to the limited context window size of GPT-4, processing all candidate nodes simultaneously may not be feasible. To address this, we employ a batch-wise approach, where we divide the candidate nodes into batches of 2,000. For each batch, we prompt GPT-4 to select the most likely candidate node. After processing all the batches, we compile the selected candidate nodes from each batch and include them as hints for the final prediction of the missing node. By providing GPT-4 with a refined set of candidate nodes, we aim to enhance its ability to accurately predict the missing node in the transductive setting.

\paragraph{Hyperparameters}
Here we provide the hyper parameters used in our experiments. All the results are provided with GPT-4-32k model with temperature 0.0 and 2000  maximum number of output tokens. For reproducibility of results we have provided all the prompts used in the experiments in the appendix.

\paragraph{Baselines}
Our first baseline involved implementing a vanilla GPT-4 prediction without any context window, where it was tasked with predicting the correct missing node (head or tail) from each triplet in the test set. Results are reported in Table \ref{tab:results_transductive}. For our second baseline, we implemented the method proposed by Chepurova et al. \cite{chepurova2023better}. Here, we augmented each triplet in the inference test set with the 1-hop neighbors of the head (or tail) entity obtained from the ILPC transductive train graph. GPT-4 then received this contextual information and was tasked with predicting the missing tail (or head) of the triplet.
Results are reported in \ref{tab:results_transductive}. For the inductive setting where the LLM did not have access to the test entities, we compared our results with two baselines reported by the authors of the ILPC small and large datasets \cite{Galkin2022}. These baselines evaluate two variants of the NodePiece model \cite{galkin2021nodepiece}, a proposed method for inductive graph representation learning. Results are reported in \ref{tab:results_inductive}

\paragraph{Evaluation Metrics}
To assess the performance of our knowledge graph completion approach, we employ the widely adopted Hit@k evaluation metric, specifically focusing on Hit@1, Hit@3, and Hit@10. The Hit@k metric measures the accuracy of the model in predicting the correct missing node within the top k ranked candidates. In our evaluation, we consider three different values of k to provide a comprehensive understanding of the model's performance at various levels of precision. Hit@1 represents the strictest evaluation criterion, where the model is considered successful only if the correct missing entity is ranked as the top candidate. This metric assesses the model's ability to accurately identify the most likely answer for a given query. Hit@3 and Hit@10 provide more relaxed evaluation criteria, allowing the correct missing node to be ranked within the top 3 and top 10 candidates, respectively.

\subsection{Results and Analysis}
In this section, we present a comprehensive analysis of our experimental results and discuss the key insights and implications derived from our findings.

{\bf{Finding 1: LLMs demonstrate strong performance in knowledge graph completion tasks.}}

Our experimental results show that GPT-4, even without any additional context or information, performs significantly better than the baselines on the ILPC-small and ILPC-large datasets. As shown in Table \ref{tab:results_transductive}, GPT-4 obtains Hit@1 scores of 0.132 and 0.146 on the ILPC-small and ILPC-large datasets, respectively. The table \ref{tab:results_transductive} further illustrates similar performance with Hit@3 and Hit@10 as well.

{\bf{Finding 2: Leveraging candidate solutions significantly improves LLM performance.}}

In the transductive setting, where test nodes are a subset of nodes in training graph, our approach of generating candidate solutions based on the ontology and utilizing LLMs to select the correct answer yields substantial performance gains. As shown in Table \ref{tab:results_transductive} GPT-4 + candidate solutions achieves Hit@1 scores of 0.172 and 0.177 on the ILPC-small and ILPC-large datasets, respectively, outperforming the GPT-4 baseline and the GPT-4 + neighbors approach proposed by \cite{chepurova2023better}. Furthermore, incorporating ontology information and ontology paths alongside the candidate solutions leads to even higher performance. However results illustrate that adding both ontology hint and ontology paths does not necessarily offer a performance improvement. The reason for this is that ontology paths already contains the information about the node category of the missing node, which is the piece of information provided in the ontology method.

{\bf{Finding 3: Incorporating ontology information enhances LLM performance in the inductive setting}}

By providing GPT-4 with ontology information, such as the category of the answer and paths between the head and tail categories, we observe a consistent improvement in performance across both the ILPC-small and ILPC-large datasets. As shown in Table \ref{tab:results_inductive}, GPT-4 + ontology achieves Hit@1 scores of 0.134 and 0.150 on the ILPC-small and ILPC-large datasets, respectively, outperforming the baseline GPT-4 model. Similarly, GPT-4 + ontology paths and GPT-4 + ontology + ontology paths demonstrate even higher performance, with Hit@1 scores reaching 0.138 and 0.152 on the ILPC-small and ILPC-large datasets, respectively. These results suggest that incorporating ontological knowledge and structural information from the graph can guide LLMs towards more accurate predictions in the inductive setting.

{\bf{Finding 4: Our approach outperforms state-of-the-art baselines in both inductive and transductive settings}}

Comparing our results to the state-of-the-art baselines, IndNodePiece and IndNodePieceGNN \cite{galkin2021nodepiece}. We observe that our approach significantly outperforms these methods in both the inductive and transductive settings. As shown in Table 2, our best-performing model, GPT-4 + ontology + ontology paths, achieves Hit@1 scores of 0.138 and 0.152 on the ILPC-small and ILPC-large datasets, respectively, in the inductive setting, surpassing the baselines by a wide margin. Similarly, in the transductive setting, our approach demonstrates superior performance, with GPT-4 + candidate solutions + ontology paths achieving Hit@1 scores of 0.174 and 0.178 on the ILPC-small and ILPC-large datasets, respectively. These results highlight the effectiveness of our approach in leveraging LLMs and incorporating topological information for knowledge graph completion. 

{\bf{Finding 5: Increasing the number of candidate nodes given in the context improves the performance.}}

Table \ref{tab:candidate_node} presents the performance of our proposed methods on the ILPC-small dataset when using different numbers of candidate nodes are given in the context window. The numbers outside the parentheses represent the case where we choose a single candidate node from each sublist of the original candidate nodes. In contrast, the numbers inside the parentheses correspond to the case where we select multiple candidate nodes from each sublist, such that the total number of chosen candidate nodes is 100. We observe a consistent improvement in performance when increasing the number of candidate nodes provided in the context. For instance, GPT-4 + candidate solutions achieves Hit@1, Hit@3, and Hit@10 scores of 0.172, 0.233, and 0.319, respectively, when using a single candidate node from each sublist. However, when using multiple candidate nodes totaling 100, the scores increase to 0.196, 0.305, and 0.391, respectively. This trend holds for all the other methods as well, demonstrating that providing a larger number of relevant candidate nodes in the context enhances the LLM's ability to accurately predict the missing node.

{\bf{Finding 6: Predicting head nodes directly yields similar performance to predicting tail nodes with inverse relations.}}

In the literature, the problem of predicting head nodes is typically converted to a tail node prediction problem by considering the inverse of the relation, denoted as $r^{-1}$ which constructed by adding the word \textit{inverse} as a prefix to the original relation. For a triplet ($v_i$, $r$, $v_j$), predicting the head node $v_i$ is transformed into predicting the tail node in ($v_j$, $r^{-1}$, ?). However, in our experiments, we compared the results of directly predicting the head node (?, $r$, $v_j$) with the inverse relation approach ($v_j$, $r^{-1}$, ?) using GPT-4 with the proposed methods. Interestingly, we found no significant difference in performance between these two approaches when using GPT-4 with our proposed methods. 
To validate this finding, we conducted experiments on the ILPC-small dataset using both the direct head node prediction and the inverse relation approach with vanilla GPT-4 prompting and ontology based prompting. The results consistently showed that the performance of direct head node prediction was comparable to that of tail node prediction with inverse relations across all methods. Table \ref{tab:direct_inverse} presents the number of correct head node predictions made by direct head node prediction method compared to using inverse relation head node prediction method out of the total 2902 head node predictions.

%% file: sec/conclusion.tex
In this paper, we have presented a novel approach to KGC leveraging LLMs in both inductive and transductive settings. Our method introduces a generative ontology creation process using LLMs, which extracts structured knowledge directly from raw knowledge graph data. This ontology serves as a foundation, providing crucial cues for inferring missing node categories and pathways between ontology entities.

In the inductive setting, our approach utilizes the ontology and category inference to enhance missing node prediction, demonstrating significant improvements in predictive accuracy without requiring additional training. Additionally, in the transductive setting,our method effectively identifies candidate solutions for triplets using the ontology and selects the correct solution through LLM inference.

These preliminary results are promising and indicate the potential of our approach. Moving forward, our research directions include incorporating important pathways between nodes, exploring online learning techniques to adapt to evolving knowledge graphs with varying triplet probabilities, integrating additional information from diverse external sources into the graph, and expanding our experimental validation to include more datasets.

%% file: sec/limitations.tex
The ontology constructed in our method operates under a closed world assumption, where no new entities are added after the initial ontology creation phase. This assumption may limit the adaptability of our approach in dynamic knowledge graph environments where new entities frequently emerge. Looking ahead, addressing this limitation will guide us in enhancing the robustness and applicability of our framework.
Another limitation of our approach is that its performance may be impacted by the density of the graph dataset. The effectiveness of leveraging ontology paths relies on the presence of a rich set of relationships and connections within the graph. In cases where the knowledge graph is sparse and lacks a sufficient number of connections between nodes, the ontology paths may not provide significant support for link prediction.

%% file: appendix/prompts.tex
\paragraph{Vannila Prompt}

\textit{You will receive a triplet with a missing node. Given triplet can be of the form available node --> relation --> ? or ? --> relation --> available node. Your task is predicting the missing node. Alongside, you will get a hint about the type of answer that is correct, and you might receive additional relevant information to aid your prediction. Please respond only with the missing node, without including the head node or the relation.}

\textit{Your task:}

\textit{Triplet with missing node:}
\textit{\{triplet\}}

\textit{Please provide a list of 10 candidate nodes for the missing node. Answer should be of the format ['candidate\_node\_1', 'candidate\_node\_2', ......, 'candidate\_node\_10']. The list of candidate nodes should be in order of most probably candidate node to least probable candidate node.}

\textit{Missing candidate nodes:}

\paragraph{Ontology Prompt}
\textit{You will receive a triplet with a missing node. Given triplet can be of the form available node --> relation --> ? or ? --> relation --> available node. Your task is predicting the missing node. Alongside, you will get a hint about the type of answer that is correct, and you might receive additional relevant information to aid your prediction. Please respond only with the missing node, without including the head node or the relation.}

\textit{Your task:}

\textit{Triplet with missing node:}
\textit{\{triplet\}}

\textit{Hint about missing node type:}
\textit{The missing node should be of type \{type\}}

\textit{Please provide a list of 10 candidate nodes for the missing node. Answer should be of the format ['candidate\_node\_1', 'candidate\_node\_2', ......, 'candidate\_node\_10']. The list of candidate nodes should be in order of most probably candidate node to least probable candidate node.}

\textit{Missing candidate nodes:}

\paragraph{Ontology Paths}
\textit{You will receive a triplet with a missing node. Given triplet can be of the form available node --> relation --> ? or ? --> relation --> available node. Your task is predicting the missing node. Alongside, you will get a hint about the type of answer that is correct, and you might receive additional relevant information to aid your prediction. Please respond only with the missing node, without including the head node or the relation.}

\textit{Example:}

\textit{Triplet with missing node:}
\textit{John Lennon --> born\_in --> ?}

\textit{Available node John Lennon is of node type person.}

\textit{Graph paths that can be important for filling missing information for a triplet type person --> born\_in --> country are [person --> died\_in --> country, person --> child\_of --> person --> citizen\_of --> country].}

\textit{Chain of thought:}
\textit{This can be a list of questions that might help predict the missing node: [In which country the person died?, If a person is a child of another person who is a citizen of a certain country, which country is that?].}

\textit{If John Lennon --> died\_in --> United Kingdom, John Lennon --> child\_of --> Alfred Lennon --> citizen\_of --> United Kingdom}
\textit{then it is likely that}
\textit{John Lennon --> born\_in  --> United Kingdom}

\textit{Missing node:}
\textit{United Kingdom}

\textit{Your task:}

\textit{Triplet with missing node:}
\textit{\{triplet\}}

\textit{Available node \{known node\} is of node type \{type\}.}

\textit{Graph paths that are important for filling missing information for a triplet type  are \{ontology paths\}.}

\textit{Reason about the missing node using Chain of Thought method.}

\textit{Please provide a list of 10 candidate nodes for the missing node. Answer should be of the format ['candidate\_node\_1', 'candidate\_node\_2', ......, 'candidate\_node\_10']. The list of candidate nodes should be in order of most probably candidate node to least probable candidate node.}

\textit{Missing candidate nodes:}
\paragraph{Ontology and Ontology paths prompt}
\textit{You will receive a triplet with a missing node. Given triplet can be of the form available node --> relation --> ? or ? --> relation --> available node. Your task is predicting the missing node. Alongside, you will get a hint about the type of answer that is correct, and you might receive additional relevant information to aid your prediction. Please respond only with the missing node, without including the head node or the relation.}

\textit{Example:}

\textit{Triplet with missing node:}
\textit{John Lennon --> born\_in --> ?}

\textit{Available node John Lennon is of node type person.}

\textit{Graph paths that can be important for filling missing information for a triplet type person --> born\_in --> country are [person --> died\_in --> country, person --> child\_of --> person --> citizen\_of --> country].}

\textit{Chain of thought:}
\textit{This can be a list of questions that might help predict the missing node: [In which country the person died?, If a person is a child of another person who is a citizen of a certain country, which country is that?].}

\textit{If John Lennon --> died\_in --> United Kingdom, John Lennon --> child\_of --> Alfred Lennon --> citizen\_of --> United Kingdom}
\textit{then it is likely that}
\textit{John Lennon --> born\_in  --> United Kingdom}

\textit{Hint about missing node type:}
\textit{The missing node should be of type country}

\textit{Missing node:}
\textit{United Kingdom}

\textit{Your task:}

\textit{Triplet with missing node:}
\textit{\{triplet\}}

\textit{Available node \{known node\} is of node type \{type\}.}

\textit{Graph paths that are important for filling missing information for a triplet type are \{ontology paths\}.}

\textit{Reason about the missing node using Chain of Thought method.}

\textit{Hint about the missing node type:}
\textit{The missing node should be of type \{type\}}

\textit{Please provide a list of 10 candidate nodes for the missing node. Answer should be of the format ['candidate\_node\_1', 'candidate\_node\_2', ......, 'candidate\_node\_10']. The list of candidate nodes should be in order of most probably candidate node to least probable candidate node.}

\textit{Missing candidate nodes:}

\paragraph{Neighbor Prompt}
\textit{You will receive a triplet with a missing node. Given triplet can be of the form available node --> relation --> ? or ? --> relation --> available node. Your task is predicting the missing node. Alongside, you will get a hint about the type of answer that is correct, and you might receive additional relevant information to aid your prediction. Please respond only with the missing node, without including the head node or the relation.}

\textit{Your task:}

\textit{Triplet with missing node:}
\textit{\{triplet\}}

\textit{1-hop neighbours of the available node \{known node\} are given along with their relations as a list:}
\textit{\{neighbours\}. This does not mean that the missing node is in this list. It is just a hint to help you predict the missing node.}

\textit{Please provide a list of 10 candidate nodes for the missing node. Answer should be of the format ['candidate\_node\_1', 'candidate\_node\_2', ......, 'candidate\_node\_10']. The list of candidate nodes should be in order of most probably candidate node to least probable candidate node.}

\textit{Missing candidate nodes:}
\paragraph{Candidate nodes prompt}
\textit{You will receive a triplet with a missing node. Given triplet can be of the form available node --> relation --> ? or ? --> relation --> available node. Your task is predicting the missing node. Alongside, you will get a hint about the type of answer that is correct, and you might receive additional relevant information to aid your prediction. Please respond only with the missing node, without including the head node or the relation.}

\textit{Your task:}

\textit{Triplet with missing node:}
\textit{\{triplet\}}

\textit{Hint about missing node:}
\textit{The missing node should be of type \{type\}. Potential candidate nodes for the missing node are \{data\}. This does not mean that missing node is always in the provided list. It is a hint to help you predict the missing node.}

\textit{Please provide a list of 10 candidate nodes for the missing node. Answer should be of the format ['candidate\_node\_1', 'candidate\_node\_2', ......, 'candidate\_node\_10']. The list of candidate nodes should be in order of most probably candidate node to least probable candidate node.}

\textit{Missing candidate nodes:}

\paragraph{Candidate nodes with Ontology hint prompt}
 Same as Candidate nodes prompt with \\
\textit{Hint about the missing node type:}
\textit{The missing node should be of type \{type\}}

\paragraph{Candidate nodes with Ontology paths prompt}
\textit{You will receive a triplet with a missing node. Given triplet can be of the form available node --> relation --> ? or ? --> relation --> available node. Your task is predicting the missing node. Alongside, you will get a hint about the type of answer that is correct, and you might receive additional relevant information to aid your prediction. Please respond only with the missing node, without including the head node or the relation.}

\textit{Example:}

\textit{Triplet with missing node:}
\textit{John Lennon --> born\_in --> ?}

\textit{Available node John Lennon is of node type person.}

\textit{Graph paths that can be important for filling missing information for a triplet type person --> born\_in --> country are [person --> died\_in --> country, person --> child\_of --> person --> citizen\_of --> country].}

\textit{Chain of thought:}
\textit{This can be a list of questions that might help predict the missing node: [In which country the person died?, If a person is a child of another person who is a citizen of a certain country, which country is that?].}

\textit{If John Lennon --> died\_in --> United Kingdom, John Lennon --> child\_of --> Alfred Lennon --> citizen\_of --> United Kingdom}
\textit{then it is likely that}
\textit{John Lennon --> born\_in  --> United Kingdom}

\textit{Hint about missing node type:}
\textit{The missing node should be of type country}

\textit{Missing node:}
\textit{United Kingdom}

\textit{Your task:}

\textit{Triplet with missing node:}
\textit{\{triplet\}}

\textit{Available node \{known node\} is of node type \{type\}.}

\textit{Hint about missing node:}
\textit{The missing node should be of type \{type\}. Potential candidate nodes for the missing node are \{data\}. This does not mean that missing node is always in the provided list. It is a hint to help you predict the missing node.}

\textit{Graph paths that are important for filling missing information for a triplet type are \{ontology paths\}.}

\textit{Reason about the missing node using Chain of Thought method.}

\textit{Please provide a list of 10 candidate nodes for the missing node. Answer should be of the format ['candidate\_node\_1', 'candidate\_node\_2', ......, 'candidate\_node\_10']. The list of candidate nodes should be in order of most probably candidate node to least probable candidate node.}

\textit{Missing candidate nodes:}

\paragraph{Candidate nodes with Ontology paths and hit prompt}
Same as Candidate nodes with Ontology paths prompt with \\
\textit{Hint about the missing node type:}
\textit{The missing node should be of type \{type\}}

\paragraph{Candidate nodes with ontology hint and graph paths prompt}
\textit{You will receive a triplet with a missing node. Given triplet can be of the form available node --> relation --> ? or ? --> relation --> available node. Your task is predicting the missing node. Alongside, you will get a hint about the type of answer that is correct, and you might receive additional relevant information to aid your prediction. Please respond only with the missing node, without including the head node or the relation.}

\textit{Example:}

\textit{Triplet with missing node:}
\textit{John Lennon --> born\_in --> ?}

\textit{Available node John Lennon is of node type person.}

\textit{Graph paths that can be important for filling missing information for triplet John Lennon --> born\_in --> ? are [John Lennon --> died\_in --> United Kingdom, John Lennon --> child\_of --> Alfred Lennon --> citizen\_of --> United Kingdom].}

\textit{Chain of thought:}

\textit{If John Lennon died\_in United Kingdom, and John Lennon is a child\_of Alfred Lennon who is a citizen\_of United Kingdom}
\textit{then it is likely that}
\textit{John Lennon --> born\_in  --> United Kingdom}

\textit{Hint about missing node type:}
\textit{The missing node should be of type country}

\textit{Missing node:}
\textit{United Kingdom}

\textit{Your task:}

\textit{Triplet with missing node:}
\textit{\{triplet\}}

\textit{Hint about missing node:}
\textit{The missing node should be of type \{type\}. Potential candidate nodes for the missing node are \{data\}. This does not mean that missing node is always in the provided list. It is a hint to help you predict the missing node.}

\textit{Graph paths that are important for filling missing information for triplet \{triplet\} are \{graph paths\}.}

\textit{Reason about the missing node using Chain of Thought method.}

\textit{Hint about the missing node type:}
\textit{The missing node should be of type \{type\}}

\textit{Please provide a list of 10 candidate nodes for the missing node. Answer should be of the format ['candidate\_node\_1', 'candidate\_node\_2', ......, 'candidate\_node\_10']. The list of candidate nodes should be in order of most probably candidate node to least probable candidate node.}

\textit{Missing candidate nodes:}

\paragraph{Ontology Prompt}
\textit{I will provide you with a relation and two data pairs. Your task is to determine the specific ontology node classes for the entities in these data pairs that are connected by the specified relation.}

\textit{Strict Requirements:}
\textit{1) All node classes must be written in lowercase.}
\textit{2) Connect words within node classes using underscores.}
\textit{3) The relation provided must not be altered in your response.}
\textit{4) Provide a single response for all data pairs, formatted as follows: ['head node class', 'tail node class', 'relation'].}
\textit{5) Ensure your answer strictly adheres to this format: ['head node class', 'tail node class', 'relation']. Do not include any additional text or explanation.}

\textit{Soft Requirements:}
\textit{1) Refer to the existing node classes: \{ontology\_categories\}. You may reuse these if they accurately describe the data pairs. If not, provide a more suitable classification.}
\textit{2) Avoid using generic terms like 'person'. Instead, use more specific classifications such as 'film\_producer' or 'play\_writer' where applicable.}

\textit{Example 1:}
\textit{Relation: 'was born in'}
\textit{Data Pairs: ['(John Lennon, United Kingdom)', '(Miles Davis, United States)']}
\textit{Answer:}
\textit{['musician', 'country', 'was born in']}

\textit{Example 2:}
\textit{Relation: 'directed by'}
\textit{Data Pairs: ['(Inception, Christopher Nolan)', '(Titanic, James Cameron)']}
\textit{Answer:}
\textit{['film', 'film\_director', 'directed by']}

\textit{Your turn:}
\textit{Relation: '\{relation\}'}
\textit{Data Pairs: '\{data\_pairs\}'}
\textit{Answer:}

%% file: appendix/ontology_paths.tex
\subsection{Alternate ontology paths for relations in ILPC-small dataset}
In this section we provide the alternate ontology paths for a given ontology triplet

{\bf{Individual}} \ra {\bf{medical condition}} \ra {\bf{medical condition}}
\setlist{nolistsep}
\begin{itemize}[noitemsep]
    \item Individual \ra cause of death \ra medical condition
\end{itemize}

{\bf{Individual}} \ra {\bf{place of birth}} \ra {\bf{city}}
\setlist{nolistsep}
\begin{itemize}[noitemsep]
    \item Individual \ra employer \ra university or organization \ra  headquarters location \ra city
    \item Individual \ra place of death \ra city
    \item Individual \ra residence \ra city
\end{itemize}

{\bf{Individual}} \ra {\bf{part of}} \ra {\bf{organization}}
\setlist{nolistsep}
\begin{itemize}[noitemsep]
    \item Individual \ra member of \ra organization
\end{itemize}

{\bf{Individual}} \ra {\bf{residence}} \ra {\bf{city}}
\setlist{nolistsep}
\begin{itemize}[noitemsep]
    \item Individual \ra employer \ra university or organization \ra  headquarters location \ra city
    \item Individual \ra place of death \ra city
    \item Individual \ra place of birth \ra city
\end{itemize}

{\bf{Individual}} \ra {\bf{languages spoken, written or signed}} \ra {\bf{language}}
\setlist{nolistsep}
\begin{itemize}[noitemsep]
    \item Individual \ra employer \ra university or organization \ra  located in the administrative territorial entity \ra city or country \ra  official language \ra language
\end{itemize}

\textbf{Individual} \ra \textbf{member of} \ra \textbf{organization}
\begin{itemize}
    \item Individual \ra part of \ra organization
\end{itemize}

\textbf{Individual} \ra \textbf{place of death} \ra \textbf{city}
\begin{itemize}
    \item Individual \ra employer \ra university or organization \ra headquarters location \ra city
    \item Individual \ra residence \ra city
    \item Individual \ra place of birth \ra city
\end{itemize}

\textbf{University or organization} \ra \textbf{headquarters location} \ra \textbf{city}
\begin{itemize}
    \item University or organization \ra located in the administrative territorial entity \ra city or country \ra  named after \ra individual \ra  place of death \ra city
    \item University or organization \ra located in the administrative territorial entity \ra city or country \ra  named after \ra individual \ra  residence \ra city
    \item University or organization \ra located in the administrative territorial entity \ra city or country \ra  named after \ra individual \ra  place of birth \ra city
\end{itemize}

\textbf{Individual} \ra \textbf{cause of death} \ra \textbf{medical condition}
\begin{itemize}
    \item Individual \ra medical condition \ra medical condition
\end{itemize}

\textbf{City or country} \ra \textbf{official language} \ra \textbf{language}
\begin{itemize}
    \item City or country \ra named after \ra individual \ra  languages spoken written or signed \ra language
\end{itemize}

\subsection{Alternate ontology paths for relations in ILPC-large dataset}

\textbf{Individual} \ra \textbf{field of work} \ra \textbf{field of work}
\begin{itemize}
    \item Individual \ra is the study of \ra field of work
\end{itemize}

\textbf{Location} \ra \textbf{country} \ra \textbf{country}
\begin{itemize}
    \item Location \ra named after \ra individual \ra country of citizenship \ra country
    \item Location \ra located in the administrative territorial entity \ra administrative territorial entity \ra legislative body \ra organization \ra founded by \ra individual \ra country of citizenship \ra country
    \item Location \ra located in the administrative territorial entity \ra administrative territorial entity \ra legislative body \ra organization \ra chief executive officer \ra individual \ra country of citizenship \ra country
    \item Location \ra located in the administrative territorial entity \ra administrative territorial entity \ra legislative body \ra organization \ra chairperson \ra individual \ra country of citizenship \ra country
\end{itemize}

\textbf{Location} \ra \textbf{has use} \ra \textbf{sport}
\begin{itemize}
    \item Location \ra named after \ra individual \ra sport \ra sport
    \item Location \ra located in the administrative territorial entity \ra administrative territorial entity \ra legislative body \ra organization \ra founded by \ra individual \ra sport \ra sport
    \item Location \ra located in the administrative territorial entity \ra administrative territorial entity \ra legislative body \ra organization \ra chief executive officer \ra individual \ra sport \ra sport
    \item Location \ra located in the administrative territorial entity \ra administrative territorial entity \ra legislative body \ra organization \ra chairperson \ra individual \ra sport \ra sport
\end{itemize}

\textbf{Individual} \ra \textbf{languages spoken, written or signed} \ra \textbf{language}
\begin{itemize}
    \item Individual \ra place of burial \ra location \ra located in the administrative territorial entity \ra administrative territorial entity \ra official language \ra language
    \item Individual \ra part of \ra organization \ra airline hub \ra location \ra located in the administrative territorial entity \ra administrative territorial entity \ra official language \ra language
\end{itemize}

\textbf{Individual} \ra \textbf{movement} \ra \textbf{genre}
\begin{itemize}
    \item Individual \ra genre \ra genre
\end{itemize}

\textbf{Individual} \ra \textbf{employer} \ra \textbf{educational institution}
\begin{itemize}
    \item Individual \ra educated at \ra educational institution
\end{itemize}

\textbf{Organization} \ra \textbf{chairperson} \ra \textbf{individual}
\begin{itemize}
    \item Organization \ra founded by \ra individual
    \item Organization \ra chief executive officer \ra individual
    \item Organization \ra airline hub \ra location \ra named after \ra individual
\end{itemize}

\textbf{Organization} \ra \textbf{chief executive officer} \ra \textbf{individual}
\begin{itemize}
    \item Organization \ra founded by \ra individual
    \item Organization \ra chairperson \ra individual
    \item Organization \ra airline hub \ra location \ra named after \ra individual
\end{itemize}

\textbf{Individual} \ra \textbf{educated at} \ra \textbf{educational institution}
\begin{itemize}
    \item Individual \ra employer \ra educational institution
\end{itemize}

\textbf{Creative work} \ra \textbf{country of origin} \ra \textbf{country}
\begin{itemize}
    \item Creative work \ra creator \ra individual \ra place of burial \ra location \ra country \ra country
    \item Creative work \ra creator \ra individual \ra country of citizenship \ra country
    \item Creative work \ra creator \ra individual \ra part of \ra organization \ra airline hub \ra location \ra country \ra country
    \item Creative work \ra publisher \ra organization \ra founded by \ra individual \ra place of burial \ra location \ra country \ra country
    \item Creative work \ra publisher \ra organization \ra founded by \ra individual \ra country of citizenship \ra country
    \item Creative work \ra publisher \ra organization \ra chief executive officer \ra individual \ra place of burial \ra location \ra country \ra country
    \item Creative work \ra publisher \ra organization \ra chief executive officer \ra individual \ra country of citizenship \ra country
    \item Creative work \ra publisher \ra organization \ra chairperson \ra individual \ra place of burial \ra location \ra country \ra country
    \item Creative work \ra publisher \ra organization \ra chairperson \ra individual \ra country of citizenship \ra country
    \item Creative work \ra publisher \ra organization \ra airline hub \ra location \ra country \ra country
    \item Creative work \ra publisher \ra organization \ra airline hub \ra location \ra named after \ra individual \ra country of citizenship \ra country
    \item Creative work \ra narrative location \ra location \ra country \ra country
    \item Creative work \ra narrative location \ra location \ra named after \ra individual \ra country of citizenship \ra country
    \item Creative work \ra narrative location \ra location \ra located in the administrative territorial entity \ra administrative territorial entity \ra legislative body \ra organization \ra founded by \ra individual \ra country of citizenship \ra country
    \item Creative work \ra narrative location \ra location \ra located in the administrative territorial entity \ra administrative territorial entity \ra legislative body \ra organization \ra chief executive officer \ra individual \ra country of citizenship \ra country
    \item Creative work \ra narrative location \ra location \ra located in the administrative territorial entity \ra administrative territorial entity \ra legislative body \ra organization \ra chairperson \ra individual \ra country of citizenship \ra country
\end{itemize}

\textbf{Organization} \ra \textbf{headquarters location} \ra \textbf{city}
\begin{itemize}
    \item Organization \ra founded by \ra individual \ra place of death \ra city
    \item Organization \ra founded by \ra individual \ra place of birth \ra city
    \item Organization \ra founded by \ra individual \ra residence \ra city
    \item Organization \ra founded by \ra individual \ra place of burial \ra location \ra located in the administrative territorial entity \ra administrative territorial entity \ra capital \ra city
    \item Organization \ra chief executive officer \ra individual \ra place of death \ra city
    \item Organization \ra chief executive officer \ra individual \ra place of birth \ra city
    \item Organization \ra chief executive officer \ra individual \ra residence \ra city
    \item Organization \ra chief executive officer \ra individual \ra place of burial \ra location \ra located in the administrative territorial entity \ra administrative territorial entity \ra capital \ra city
    \item Organization \ra chairperson \ra individual \ra place of death \ra city
    \item Organization \ra chairperson \ra individual \ra place of birth \ra city
    \item Organization \ra chairperson \ra individual \ra residence \ra city
    \item Organization \ra chairperson \ra individual \ra place of burial \ra location \ra located in the administrative territorial entity \ra administrative territorial entity \ra capital \ra city
    \item Organization \ra airline hub \ra location \ra named after \ra individual \ra place of death \ra city
    \item Organization \ra airline hub \ra location \ra named after \ra individual \ra place of birth \ra city
    \item Organization \ra airline hub \ra location \ra named after \ra individual \ra residence \ra city
    \item Organization \ra airline hub \ra location \ra located in the administrative territorial entity \ra administrative territorial entity \ra capital \ra city
\end{itemize}

\textbf{Individual} \ra \textbf{residence} \ra \textbf{city}
\begin{itemize}
    \item Individual \ra place of death \ra city
    \item Individual \ra place of birth \ra city
    \item Individual \ra place of burial \ra location \ra located in the administrative territorial entity \ra administrative territorial entity \ra capital \ra city
    \item Individual \ra place of burial \ra location \ra located in the administrative territorial entity \ra administrative territorial entity \ra legislative body \ra organization \ra headquarters location \ra city
    \item Individual \ra part of \ra organization \ra headquarters location \ra city
    \item Individual \ra part of \ra organization \ra airline hub \ra location \ra located in the administrative territorial entity \ra administrative territorial entity \ra capital \ra city
\end{itemize}

\textbf{Individual} \ra \textbf{place of birth} \ra \textbf{city}
\begin{itemize}
    \item Individual \ra place of death \ra city
    \item Individual \ra residence \ra city
    \item Individual \ra place of burial \ra location \ra located in the administrative territorial entity \ra administrative territorial entity \ra capital \ra city
    \item Individual \ra place of burial \ra location \ra located in the administrative territorial entity \ra administrative territorial entity \ra legislative body \ra organization \ra headquarters location \ra city
    \item Individual \ra part of \ra organization \ra headquarters location \ra city
    \item Individual \ra part of \ra organization \ra airline hub \ra location \ra located in the administrative territorial entity \ra administrative territorial entity \ra capital \ra city
\end{itemize}

\textbf{Creative work} \ra \textbf{narrative location} \ra \textbf{location}
\begin{itemize}
    \item Creative work \ra creator \ra individual \ra place of burial \ra location
    \item Creative work \ra creator \ra individual \ra part of \ra organization \ra airline hub \ra location
    \item Creative work \ra publisher \ra organization \ra founded by \ra individual \ra place of burial \ra location
    \item Creative work \ra publisher \ra organization \ra chief executive officer \ra individual \ra place of burial \ra location
    \item Creative work \ra publisher \ra organization \ra chairperson \ra individual \ra place of burial \ra location
    \item Creative work \ra publisher \ra organization \ra airline hub \ra location
\end{itemize}

\textbf{Individual} \ra \textbf{place of burial} \ra \textbf{location}
\begin{itemize}
    \item Individual \ra part of \ra organization \ra airline hub \ra location
\end{itemize}

\textbf{Individual} \ra \textbf{country of citizenship} \ra \textbf{country}
\begin{itemize}
    \item Individual \ra place of burial \ra location \ra country \ra country
    \item Individual \ra part of \ra organization \ra airline hub \ra location \ra country \ra country
\end{itemize}

\textbf{Creative work} \ra \textbf{publisher} \ra \textbf{organization}
\begin{itemize}
    \item Creative work \ra creator \ra individual \ra place of burial \ra location \ra located in the administrative territorial entity \ra administrative territorial entity \ra legislative body \ra organization
    \item Creative work \ra creator \ra individual \ra part of \ra organization
    \item Creative work \ra narrative location \ra location \ra named after \ra individual \ra part of \ra organization
    \item Creative work \ra narrative location \ra location \ra located in the administrative territorial entity \ra administrative territorial entity \ra legislative body \ra organization
\end{itemize}

\textbf{Administrative territorial entity} \ra \textbf{capital} \ra \textbf{city}
\begin{itemize}
    \item Administrative territorial entity \ra legislative body \ra organization \ra headquarters location \ra city
    \item Administrative territorial entity \ra legislative body \ra organization \ra founded by \ra individual \ra place of death \ra city
    \item Administrative territorial entity \ra legislative body \ra organization \ra founded by \ra individual \ra place of birth \ra city
    \item Administrative territorial entity \ra legislative body \ra organization \ra founded by \ra individual \ra residence \ra city
    \item Administrative territorial entity \ra legislative body \ra organization \ra chief executive officer \ra individual \ra place of death \ra city
    \item Administrative territorial entity \ra legislative body \ra organization \ra chief executive officer \ra individual \ra place of birth \ra city
    \item Administrative territorial entity \ra legislative body \ra organization \ra chief executive officer \ra individual \ra residence \ra city
    \item Administrative territorial entity \ra legislative body \ra organization \ra chairperson \ra individual \ra place of death \ra city
    \item Administrative territorial entity \ra legislative body \ra organization \ra chairperson \ra individual \ra place of birth \ra city
    \item Administrative territorial entity \ra legislative body \ra organization \ra chairperson \ra individual \ra residence \ra city
    \item Administrative territorial entity \ra legislative body \ra organization \ra airline hub \ra location \ra named after \ra individual \ra place of death \ra city
    \item Administrative territorial entity \ra legislative body \ra organization \ra airline hub \ra location \ra named after \ra individual \ra place of birth \ra city
    \item Administrative territorial entity \ra legislative body \ra organization \ra airline hub \ra location \ra named after \ra individual \ra residence \ra city
\end{itemize}

\textbf{Location} \ra \textbf{named after} \ra \textbf{individual}
\begin{itemize}
    \item Location \ra located in the administrative territorial entity \ra administrative territorial entity \ra legislative body \ra organization \ra founded by \ra individual
    \item Location \ra located in the administrative territorial entity \ra administrative territorial entity \ra legislative body \ra organization \ra chief executive officer \ra individual
    \item Location \ra located in the administrative territorial entity \ra administrative territorial entity \ra legislative body \ra organization \ra chairperson \ra individual
\end{itemize}

\textbf{Organization} \ra \textbf{founded by} \ra \textbf{individual}
\begin{itemize}
    \item Organization \ra chief executive officer \ra individual
    \item Organization \ra chairperson \ra individual
    \item Organization \ra airline hub \ra location \ra named after \ra individual
\end{itemize}

\textbf{Organization} \ra \textbf{airline hub} \ra \textbf{location}
\begin{itemize}
    \item Organization \ra founded by \ra individual \ra place of burial \ra location
    \item Organization \ra chief executive officer \ra individual \ra place of burial \ra location
    \item Organization \ra chairperson \ra individual \ra place of burial \ra location
\end{itemize}

\textbf{Individual} \ra \textbf{genre} \ra \textbf{genre}
\begin{itemize}
    \item Individual \ra movement \ra genre
\end{itemize}

\textbf{Administrative territorial entity} \ra \textbf{official language} \ra \textbf{language}
\begin{itemize}
    \item Administrative territorial entity \ra legislative body \ra organization \ra founded by \ra individual \ra languages spoken, written or signed \ra language
    \item Administrative territorial entity \ra legislative body \ra organization \ra chief executive officer \ra individual \ra languages spoken, written or signed \ra language
    \item Administrative territorial entity \ra legislative body \ra organization \ra chairperson \ra individual \ra languages spoken, written or signed \ra language
    \item Administrative territorial entity \ra legislative body \ra organization \ra airline hub \ra location \ra named after \ra individual \ra languages spoken, written or signed \ra language
\end{itemize}

\textbf{Individual} \ra \textbf{sport} \ra \textbf{sport}
\begin{itemize}
    \item Individual \ra place of burial \ra location \ra has use \ra sport
    \item Individual \ra part of \ra organization \ra airline hub \ra location \ra has use \ra sport
\end{itemize}

\textbf{Individual} \ra \textbf{unmarried partner} \ra \textbf{individual}

\textbf{Creative work} \ra \textbf{creator} \ra \textbf{individual}
\begin{itemize}
    \item Creative work \ra publisher \ra organization \ra founded by \ra individual
    \item Creative work \ra publisher \ra organization \ra chief executive officer \ra individual
    \item Creative work \ra publisher \ra organization \ra chairperson \ra individual
    \item Creative work \ra publisher \ra organization \ra airline hub \ra location \ra named after \ra individual
    \item Creative work \ra narrative location \ra location \ra named after \ra individual
    \item Creative work \ra narrative location \ra location \ra located in the administrative territorial entity \ra administrative territorial entity \ra legislative body \ra organization \ra founded by \ra individual
    \item Creative work \ra narrative location \ra location \ra located in the administrative territorial entity \ra administrative territorial entity \ra legislative body \ra organization \ra chief executive officer \ra individual
    \item Creative work \ra narrative location \ra location \ra located in the administrative territorial entity \ra administrative territorial entity \ra legislative body \ra organization \ra chairperson \ra individual
\end{itemize}

\textbf{Individual} \ra \textbf{occupation} \ra \textbf{occupation}
\begin{itemize}
    \item Individual \ra n/a \ra occupation
\end{itemize}

\textbf{Individual} \ra \textbf{is the study of} \ra \textbf{field of work}
\begin{itemize}
    \item Individual \ra field of work \ra field of work
\end{itemize}

\textbf{Individual} \ra \textbf{place of death} \ra \textbf{city}
\begin{itemize}
    \item Individual \ra place of birth \ra city
    \item Individual \ra residence \ra city
    \item Individual \ra place of burial \ra location \ra located in the administrative territorial entity \ra administrative territorial entity \ra capital \ra city
    \item Individual \ra place of burial \ra location \ra located in the administrative territorial entity \ra administrative territorial entity \ra legislative body \ra organization \ra headquarters location \ra city
    \item Individual \ra part of \ra organization \ra headquarters location \ra city
    \item Individual \ra part of \ra organization \ra airline hub \ra location \ra located in the administrative territorial entity \ra administrative territorial entity \ra capital \ra city
\end{itemize}

\textbf{Creative work} \ra \textbf{language of work or name} \ra \textbf{language}
\begin{itemize}
    \item Creative work \ra creator \ra individual \ra place of burial \ra location \ra located in the administrative territorial entity \ra administrative territorial entity \ra official language \ra language
    \item Creative work \ra creator \ra individual \ra languages spoken, written or signed \ra language
    \item Creative work \ra creator \ra individual \ra part of \ra organization \ra airline hub \ra location \ra located in the administrative territorial entity \ra administrative territorial entity \ra official language \ra language
    \item Creative work \ra publisher \ra organization \ra founded by \ra individual \ra place of burial \ra location \ra located in the administrative territorial entity \ra administrative territorial entity \ra official language \ra language
    \item Creative work \ra publisher \ra organization \ra founded by \ra individual \ra languages spoken, written or signed \ra language
    \item Creative work \ra publisher \ra organization \ra chief executive officer \ra individual \ra place of burial \ra location \ra located in the administrative territorial entity \ra administrative territorial entity \ra official language \ra language
    \item Creative work \ra publisher \ra organization \ra chief executive officer \ra individual \ra languages spoken, written or signed \ra language
    \item Creative work \ra publisher \ra organization \ra chairperson \ra individual \ra place of burial \ra location \ra located in the administrative territorial entity \ra administrative territorial entity \ra official language \ra language
    \item Creative work \ra publisher \ra organization \ra chairperson \ra individual \ra languages spoken, written or signed \ra language
    \item Creative work \ra publisher \ra organization \ra airline hub \ra location \ra named after \ra individual \ra languages spoken, written or signed \ra language
    \item Creative work \ra publisher \ra organization \ra airline hub \ra location \ra located in the administrative territorial entity \ra administrative territorial entity \ra official language \ra language
    \item Creative work \ra narrative location \ra location \ra named after \ra individual \ra languages spoken, written or signed \ra language
    \item Creative work \ra narrative location \ra location \ra located in the administrative territorial entity \ra administrative territorial entity \ra official language \ra language
    \item Creative work \ra narrative location \ra location \ra located in the administrative territorial entity \ra administrative territorial entity \ra legislative body \ra organization \ra founded by \ra individual \ra languages spoken, written or signed \ra language
    \item Creative work \ra narrative location \ra location \ra located in the administrative territorial entity \ra administrative territorial entity \ra legislative body \ra organization \ra chief executive officer \ra individual \ra languages spoken, written or signed \ra language
    \item Creative work \ra narrative location \ra location \ra located in the administrative territorial entity \ra administrative territorial entity \ra legislative body \ra organization \ra chairperson \ra individual \ra languages spoken, written or signed \ra language
\end{itemize}

\textbf{Individual} \ra \textbf{part of} \ra \textbf{organization}
\begin{itemize}
    \item Individual \ra place of burial \ra location \ra located in the administrative territorial entity \ra administrative territorial entity \ra legislative body \ra organization
\end{itemize}

\textbf{Individual} \ra \textbf{n/a} \ra \textbf{occupation}
\begin{itemize}
    \item Individual \ra occupation \ra occupation
\end{itemize}